\documentclass[manuscript]{acmart}
\AtBeginDocument{%
  \providecommand\BibTeX{{%
    \normalfont B\kern-0.5em{\scshape i\kern-0.25em b}\kern-0.8em\TeX}}}

\copyrightyear{2023}
\acmYear{2023}
\setcopyright{rightsretained}
\acmConference[ASSETS '23]{The 25th International ACM SIGACCESS Conference
on Computers and Accessibility}{October 22--25, 2023}{New York, NY, USA}
\acmBooktitle{The 25th International ACM SIGACCESS Conference on Computers
and Accessibility (ASSETS '23), October 22--25, 2023, New York, NY, USA}
\acmDOI{10.1145/3597638.3608408}
\acmISBN{979-8-4007-0220-4/23/10}

\acmConference[ASSETS '23]{The 25th International ACM SIGACCESS Conference on Computers and Accessibility}{October 22--25,2023}{New York, NY}
\acmBooktitle{ASSETS '23: ACM SIGACCESS Conference on Computers and Accessibility,
 October 22--25, 2023, New York, NY} 

\usepackage{color, colortbl}
\definecolor{Gray}{gray}{0.93}

\usepackage{multirow}
\usepackage{multicol}
\usepackage{xspace}

\begin{document}

\title{The Sem-Lex Benchmark: Modeling ASL Signs and Their Phonemes}

\author{Lee Kezar}
\email{lkezar@usc.edu}
\affiliation{
    \institution{University of Southern California}
    \city{Los Angeles}
    \state{California}
    \country{USA}
}

\author{Elana Pontecorvo}
\email{elanajp@bu.edu } 
\affiliation{
    \institution{Boston University} 
    \city{Boston}
    \state{MA}
    \country{USA}
}

\author{Adele Daniels}
\email{adeledan@bu.edu  } 
\affiliation{
    \institution{Boston University} 
    \city{Boston}
    \state{MA}
    \country{USA}
}

\author{Connor Baer}
\email{cab9@bu.edu  } 
\affiliation{
    \institution{Boston University} 
    \city{Boston}
    \state{MA}
    \country{USA}
}

\author{Ruth Ferster}
\email{rferst@bu.edu  } 
\affiliation{
    \institution{Boston University} 
    \city{Boston}
    \state{MA}
    \country{USA}
}

\author{Lauren Berger}
\email{Lauren.berger@dell.com} 
\affiliation{
    \institution{Boston University} 
    \city{Boston}
    \state{MA}
    \country{USA}
}

\author{Jesse Thomason}
\email{jessetho@usc.edu}
\affiliation{
    \institution{University of Southern California}
    \city{Los Angeles}
    \state{California}
    \country{USA}
}

\author{Zed Sevcikova Sehyr}
\email{sehyr@chapman.edu}
\affiliation{
    \institution{Chapman University}
    \city{Irvine}
    \state{California}
    \country{USA}
}
\author{Naomi Caselli}
\email{nkc@bu.edu}
\affiliation{%
  \institution{Boston University}
  \city{Boston}
  \state{Massachusetts}
  \country{USA}
}

\renewcommand{\shortauthors}{Kezar et al.}
\newcommand\todo[1]{\textcolor{red}{#1}}

\newcommand{\name}{Sem-Lex Benchmark\xspace}
\newcommand{\sltrain}{Sem-Lex$_{train}$\xspace}
\newcommand{\slval}{Sem-Lex$_{val}$\xspace}
\newcommand{\sltest}{Sem-Lex$_{test}$\xspace}

\begin{abstract}
Sign language recognition and translation technologies have the potential to increase access and inclusion of deaf signing communities, but research progress is bottlenecked by a lack of representative data.
We introduce a new resource for American Sign Language (ASL) modeling, the \name.
The Benchmark is the current largest of its kind, consisting of over 84k videos of isolated sign productions from deaf ASL signers who gave informed consent and received compensation.
Human experts aligned these videos with other sign language resources including ASL-LEX, SignBank, and ASL Citizen, enabling useful expansions for sign and phonological feature recognition.
We present a suite of experiments which make use of the linguistic information in ASL-LEX, evaluating the practicality and fairness of the \name for isolated sign recognition (ISR).
We use an SL-GCN model to show that the phonological features are recognizable with 85\% accuracy, and that they are effective as an auxiliary target to ISR.
Learning to recognize phonological features alongside gloss results in a 6\% improvement for few-shot ISR accuracy and a 2\% improvement for ISR accuracy overall.
Instructions for downloading the data can be found at \url{https://github.com/leekezar/SemLex}.
\end{abstract}



\received{3 May 2023}

\maketitle

\section{Introduction}
Word recognition is the foundation of many automatic speech-based technologies, like voice assistants, language learning apps, and translators.
While immensely practical in day-to-day use, these technologies exclude signed languages and are inaccessible to deaf people\footnote{There have been various conventions for referring to deaf communities, but there is not broad consensus on a preferred term \cite{pudans2019deaf}. We use 'deaf' rather than other terms that are widely viewed as offensive (e.g., 'hearing impaired'). We use the lower case 'deaf' here---as opposed to the capitalized 'Deaf'---to be inclusive of people with varying auditory access and with varying identities with respect to Deaf culture.} who primarily use sign language to communicate.
There has been an increasing enthusiasm among experts in many fields, including human-computer interaction, computer vision, natural language processing, and computer graphics in developing technology for automatically understanding, processing, translating, and generating sign languages \cite{bragg2019sign, includingsl}.

However, such work has had variable levels of utility and success.
One barrier to progress is a lack of adequate sign language data. 
While an array of tasks, models, and learning procedures have been developed to focus on signed languages \cite{includingsl}, less attention has been given to building large-scale, systematically-annotated, and ethically-sourced datasets to fully realize the potential of these methods \cite{bragg2021fate}. 
Another barrier to progress is the lack of linguistically-informed approaches to sign recognition.
Most prior work has treated sign recognition as a vision problem rather than a language problem, meaning these works have little-to-no acknowledgement of structural linguistic complexities of signs. 
For example, recent evidence has shown that models which treat signs as a collection of linguistic components (rather than holistic gestures) are up to 6\% more accurate at isolated sign recognition accuracy \citep{eacl}.
In this paper, we introduce new data for the purpose of overcoming these barriers, replicating the finding that phonology improves sign recognition, and investigating other benefits, namely, few-shot generalizability and sensitivity to race and gender.

Although datasets of isolated signs have many potential uses, we position this benchmark as uniquely helpful for isolated sign recognition (ISR\footnote{The term \textit{isolated sign language recognition} or ISLR is also common. We prefer ISR to more clearly disambiguate the task from sign language identification, where a model must recognize which signed language is found in a video.}). The benchmark contains over 84k videos of isolated sign productions from deaf ASL signers who gave informed consent and received compensation. The signs were reviewed and annotated by human experts using a novel labelling system that enables rapid, reliable labelling of sign language data. The annotations are cross-referenced with reference signs from the ASL-LEX database ~\cite{sehyr2021asl, caselli2017asl}, as well as SignBank \cite{hochgesang2019asl},  and ASL Citizen \cite{desai2023asl}.
Second, we conduct a suite of experiments related to sign and phonological feature recognition.
These experiments show that incorporating linguistic information about the composition of signs, namely the phonological features extracted from ASL-LEX, enables accurate phonological feature recognition and more accurate ISR.
We also conduct a quantitative analysis of model sensitivity to signer appearance and demographics and explore the models' ability to recognize signs that had few instances in training. 

\section{Background and Related Work}
\label{rw}
Deaf communities have worked hard for the recognition of sign languages as legitimate languages, as opposed to simplistic gestural systems or manual ways of expressing spoken language.
There are ongoing campaigns in many countries around the world for legal recognition of national sign languages \cite{de2019legal}.
According to the World Federation of the Deaf (WFD), the lack of recognition, acceptance, and use of sign language represents the major barrier that prevents deaf people from accessing basic human rights, especially in developing countries \cite{wfd}.
The Linguistic Society of America passed a resolution \cite{lsa} acknowledging that sign languages are, in fact, languages with all the linguistic structure inherent to any language (syntax, morphology, phonology, prosody, etc.).
Systemic recognition of languages is important because access to sign language can be precarious.
Deaf children are often denied the opportunity to acquire a signed language putting them at risk of language deprivation during the critical window of childhood development \cite{hall2019deaf, hecht2020responsibility}.
Without recognition of sign languages and robust systems for sign language interpreting services, deaf people are often denied full access to basic aspects of life such as employment, education or healthcare \cite{xiaoyan2009survey, ball2017history}. 

Along these lines, deaf communities have  raised concerns about lack of recognition of sign languages as real languages in the development of sign language technology. 
For example, in a paper in Nature Electronics, Hill laments a ``lack of an appropriate linguistic framework'' and the ``lack of interdisciplinary collaboration'' \cite{hill2020deaf}. 
These calls highlight the need for technologists to honor sign languages as equally structured, complex, and organically-evolving as spoken languages. 
For our part, the \name is the result of collaboration among computer scientists and linguists, and directly relies on contemporary ideas in ASL phonology and machine learning.

\subsection{Insights From Research On Sign Language Phonology}
Spoken words are composed of discrete, recombinable sound units, such as vowels or consonants (phonemes), and there is a general consensus that signs are made up of a finite number of analogous phonological parameters. Early work on sign languages identified the central parameters as handshape, movement, place of articulation (location) and non-manual markers \cite{StokoeWilliamC1960Slsa}. More recent work goes beyond these basic parameters, noting that the parameters can be further described in terms of phonological features\footnote{We refer to the component parts of signs as `phonological features' rather than `phonemes'. Spoken phonemes are sequenced, discrete bundles of phonological features like voicing, place of articulation, and manner. For many signs, there is one and only one of each phonological feature (e.g., signs must have a major location, and cannot have more than one major location), and the timing and sequence of features is not segmental as it is in speech.} that have complex dependencies (e.g., handshape may be further specified in terms of selected fingers that vary in flexion and spread) \cite{van2002phonological, brentari1998prosodic, sandler1987sequentiality}. Some of these features change during the sign (e.g., the \textit{flexion} or \textit{spread} of the fingers) and some do not (e.g., the \textit{major location} of the hand, the \textit{selected fingers}). 
The study of sign language phonology is crucial for our understanding of how people learn, recognize, and produce signs.
Additionally, we find it can contribute to automatic sign recognition. 

\begin{table}[t!]
    \centering
    \begin{tabular}{llrl} 
        \toprule
        \multicolumn{1}{c}{Phonological Feature}    & \multicolumn{1}{c}{Description} & \#Values & Top Value \\
        \midrule
        Major Location                      & The broad location where the sign is produced. & 5 & /\texttt{neutral}/ \\
        
        \rowcolor{Gray} Minor Location      & The specific location where the sign is produced. & 37 & /\texttt{neutral}/ \\
        
        Second Minor Location               & The specific location after the first minor location. & 37 & /\texttt{n/a}/ \\
        
        \rowcolor{Gray} Contact             & Whether the dominant hand touches the body. & 2 & /\texttt{true}/\\
        
        Thumb Contact                       & Whether the dominant thumb touches the selected fingers. & 3 & /\texttt{false}/ \\
        
        \rowcolor{Gray}Thumb Position       & Whether the thumb is on the palm or extended. & 2 & /\texttt{open}/ \\
        
        Nondominant Handshape               & Configuration of the nondominant hand. & 56 & /\texttt{n/a}/ \\
        
        \rowcolor{Gray} Handshape           & Configuration of the dominant hand. & 58 & /\texttt{open b}/ \\
        
        Selected Fingers    & The fingers that move, or are in marked configurations. & 8 & /\texttt{imrp}/\\
        
        \rowcolor{Gray} Flexion             & The way the finger joints are bent. & 8 & /\texttt{fully open}/ \\
        
        Spread                              & Whether the selected fingers touch one another. & 3 & /\texttt{n/a}/\\
        
        \rowcolor{Gray} Spread Change       & Whether \textit{Spread} changes. & 3 & /\texttt{n/a}/ \\
        
        Repeated Movement                   & Whether the movement is repeated 2+ times. & 2 & /\texttt{false}/ \\

        \rowcolor{Gray}Sign Type            & Number of hands, and  symmetry (if two handed) & 6 & /\texttt{one handed}/ \\

        Wrist Twist                         & Whether the hand rotates about the wrist. & 2 & /\texttt{false}/ \\
        
        \rowcolor{Gray} Path Movement       & The shape that the hand traces. & 8 & /\texttt{straight}/ \\
        \bottomrule
    \end{tabular}
    \caption{Overview of each phonological feature types found in ASL-LEX, including the number of possible values and the most frequent value for each type. \texttt{n/a} appears in some Boolean phonological feature types, resulting in three possible values instead of two. \texttt{imrp} refers to \textit{index}, \textit{middle}, \textit{ring}, and \textit{pinky}. Detailed descriptions of each feature in ASL-LEX can be found in \cite{sehyr2021asl}.} 
    \label{rw:tab:ptype}
\end{table}

\subsection{Labelling and Annotating Signs}
In the absence of a standard writing system for signed languages, the question of how to best represent signing is surrounded with much debate \cite{fenlon2015building, schuller2021lemma, mesch2015gloss, hochgesang2018building, johnston1999defining}.
For the purposes of ISR, a useful labelling system should be both efficient to apply and reliably lemmatizes signs, that is, the system should produce the same label for different instances of the same sign, and different labels for signs that are distinct.

While most researchers have used English-like glosses, some signs have multiple possible English translations (one-to-many), some English words have many possible ASL translations (many-to-one), and some signs have no equivalent English translations.
Meanwhile, efforts to replace or augment English glosses with phonological information, like SignStream \cite{Neidle2018NEWS} and HamNoSys \cite{hanke2004hamnosys} rely on idiosyncratic labelling systems which require some amount of training to apply consistently and may result in different productions of the same sign to receive different labels.

Taking these considerations into account, we chose to label the videos in Sem-Lex from a large collection of reference signs.
This feature minimizes both English interference and the amount of linguistic knowledge needed for labelling.

\subsection{Existing Datasets}
There are a handful of existing datasets of isolated signs in ASL that have been used in ISR (see Table \ref{table:datasets}).
Some of these datasets were `curated', meaning they were collected from participants who were recruited to contribute data in a specific fashion, e.g., by modeling signs based on a dictionary.
Some datasets were scraped from the internet in ways that are legally and ethically questionable, often without attribution to the video creators and without informed consent of the people in the videos \cite{li2020word,joze2018ms}.
Further, some datasets include signers with unknown backgrounds---people who may or may not have lived experience of deafness and may have learned sign language as adults \cite{li2020word,joze2018ms}.
Like all languages, people who learned sign language later in life, perhaps as a second or additional language, have highly variable levels of proficiency and articulate signs differently compared to those who acquired sign language in childhood and use it as a primary language of communication \cite{marshall2021signed}.
This difference leads to heterogeneity and inconsistencies in how signs are articulated \cite{hilger2015second}.
Generally, training data should match the anticipated end user.
In most cases, the imagined end users of sign language technology are deaf signers.
Training data that consist of a broad diversity of signers, including novice signers, may be suitable for some applications and end users.
However, it is not clear that models developed on novice signers will generalize to deaf signers.
Thus, we present the \name to solve many of the issues associated with existing datasets--a curated, larger than the state-of-the-art benchmark of isolated ASL signs produced by deaf fluent signers who provided informed consent and compensated for their effort.

\begin{center}
\begin{table}[t!]
\begin{tabular}{lrrlll}
\toprule
Dataset & Number of Signs 
& Number of Videos & Source & Participants & Informed Consent\\
\midrule
Purdue RVL-SLLL \cite{martinez2002purdue} & 39 & 546 & Curated & Deaf & Yes\\
Boston ASLLVD \cite{athitsos2008american} & 2,742 & 9,794 & Curated & Deaf & Yes \\
RWTH-BOSTON-50 \cite{zahedi2005combination} & 50 & 483  & Curated & Deaf & Yes \\
MS-ASL \cite{joze2018ms} & 1,000 & 25,513  & Scraped & Unknown & No\\
WL-ASL \cite{li2020word} & 2,000 & 21,083   & Scraped & Unknown & No \\
ASL Citizen \cite{desai2023asl} & 2,731 & 83,912   & Curated & Deaf & Yes\\

 \hline
\name &  3,149  & 84,568*   & Curated & Deaf & Yes\\
\bottomrule
\end{tabular}
\caption{Existing datasets of isolated signs in ASL. *Includes unlabeled videos. 65,935 are labeled with a gloss.}
\label{table:datasets}
\end{table}
\end{center}




\section{Sem-Lex Benchmark}
\begin{figure}
    \centering
    \includegraphics[width = \textwidth]{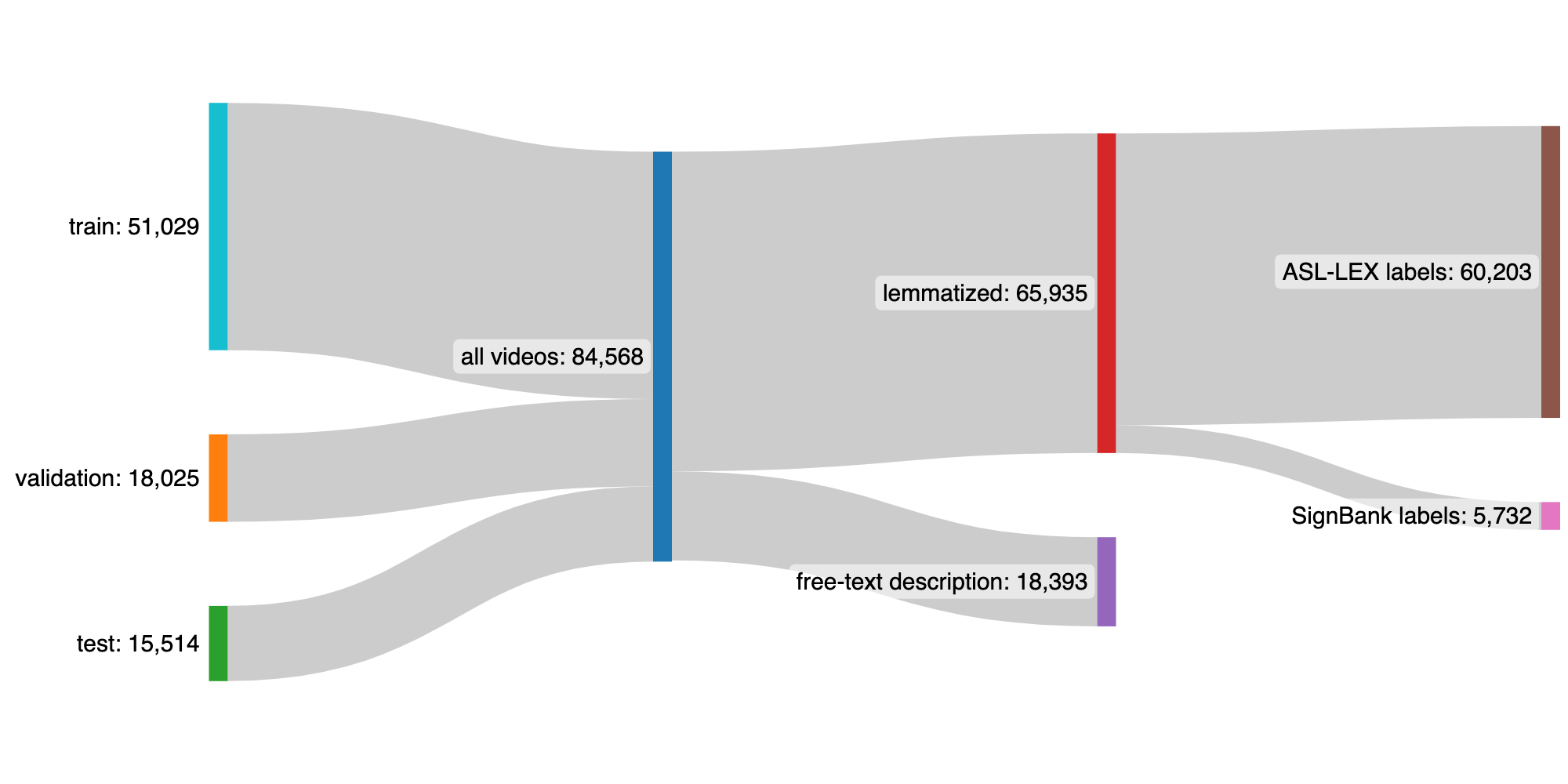}
    \caption{The \name data is divided into 3:1:1 train/validation/test, where each subset is in turn a mix of lemmatized (i.e. has been matched to an entry in a lexical database) or ``unlabeled'' (i.e. free-text description). In our experiments, we only use the lemmatized items from ASL-LEX 2.0.}
    \label{fig:sankey}
\end{figure}

The \name contributes 84,568 isolated sign videos, divided into train/validation/test splits and lemmatized ($n=65,935$) or described with free text ($n=18,393$).
Lemmatized signs were aligned with either ASL-LEX ($n=60,203$) or SignBank ($n=5,732$) (see Figure \ref{fig:sankey}).

\begin{figure}
    \centering
    \includegraphics [width = \textwidth]{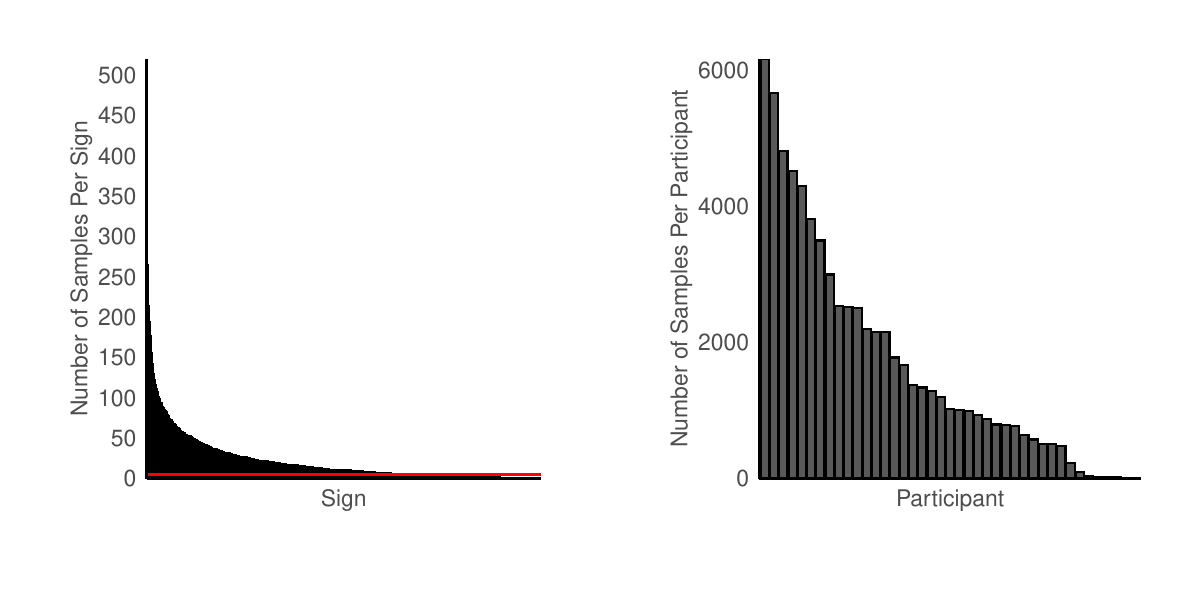}
    \caption{The distribution of samples per sign and per participant. The red line in the left panel represents 5 samples.}
    \label{fig:SamplesPerParticipant}
\end{figure}

The distribution of samples contributed by each participant is in Figure \ref{fig:SamplesPerParticipant}.
The median number of samples per sign was 10 (IQR 4-26).
A total of 3,149 unique signs were represented in the lemmatized data.
Of these, 945 signs had fewer than five samples.
To put these numbers in some perspective, the current most popular benchmark for ISR is Word-Level American Sign Language (WLASL, \cite{wlasl}), containing 21,083 videos representing 2,000 signs for an average of 10.5 video examples per sign.

\begin{itemize}
    \item \textbf{Phonological Feature Annotations.} Although all videos have a split, in this work we only use the videos which have been aligned with ASL-LEX in order to maintain consistency among the target gloss labels and complete coverage of phonological feature annotations. Future work might consider including the non-ASL-LEX videos.
    \item \textbf{Sufficient Examples.} Signs with fewer than $5$ instances are not given a split (but may be included in future work on few-shot generalizability).
    \item \textbf{Diverse, Unseen Test Set.} The test set is entirely comprised of participants who are not frequently represented in sign language training data, in order to help quantify model bias with regard to race and gender. We select 10 participants among the 41 contributors whose videos make up approximately $20\%$ of the entire dataset such that the ratio of non-white and women signers is substantially higher than average. We then place all of these participants' productions in the test set, to ensure that they are unseen during both training and validation.
\end{itemize}

\subsection{Data Collection}

The dataset consists of ASL signs elicited using a free semantic associations paradigm as part of another study aimed at understanding the lexical-semantic properties of the ASL lexicon \cite{semassoc}.
For this study, we developed an interface for rapid data collection and annotation of signs called SignLab \footnote{SignLab is a work in progress, and will be forthcoming.}.
Participants contributed data remotely from their own computers.
We asked that they ensure no other people were visible on camera, but otherwise did not control the filming conditions.
SignLab first presented participants with a video of a cue sign from ASL-LEX (e.g., CAT) and prompted them to produce the first three meaning-related signs that came to mind (e.g., DOG, MOUSE, MILK). 
Participants contributed the first three signs that came to mind by 1) pressing the space bar to turn on their webcam, 2) producing a sign, 3) pressing the space bar to turn off their camera and then repeating the process up to three times.
Participants could delete any of these responses with one button press (e.g., if there was an error), but could not re-record them.
This process enabled us to rapidly collect and segment videos so each video contained just one sign.
Because the protocol allowed participants to freely produce a sign that came to mind, it also ensured that participants knew and used each sign (i.e., rather than copying a sign they may or may not be familiar with).

Forty-one deaf ASL signers contributed data (see Table \ref{table:demographics}). Participants were paid \$15 for the initial training, \$20 per 100 trials (i.e., 100 cue signs), and a completion bonus of \$100 for every 1,000 trials they completed. All participants gave informed consent to sharing their video data in a public online repository. Consent forms were provided online in both written English and as ASL videos. Data from three participants were removed from the dataset prior to analysis because an early review of their responses indicated that they did not understand the task as intended (e.g., repeating the prompt sign, producing multi-sign responses, producing unrecognizable signs). 

\begin{table}
\small
\begin{tabular}[t]{ll}
\toprule
  & Overall\\
\midrule
 & (N=41)\\
\addlinespace[0.3em]
\multicolumn{2}{l}{\textbf{Age}}\\
\hspace{1em}Mean (SD) & 31.9 (11.6)\\
\hspace{1em}Median [Min, Max] & 27.0 [21.0, 65.0]\\
\hspace{1em}Missing & 2 \vphantom{1} (4.9\%)\\
\addlinespace[0.3em]
\multicolumn{2}{l}{\textbf{Age of First ASL Exposure}}\\
\hspace{1em}Mean (SD) & 2.00 (3.88)\\
\hspace{1em}Median [Min, Max] & 0 [0, 14.0]\\
\hspace{1em}Missing & 4 (9.8\%)\\
\addlinespace[0.3em]
\multicolumn{2}{l}{\textbf{Sex}}\\
\hspace{1em}Female & 27 (65.9\%)\\
\hspace{1em}Male & 12 (29.3\%)\\
\hspace{1em}Non Binary & 1 (2.4\%)\\
\hspace{1em}Missing & 1 \vphantom{1} (2.4\%)\\
\addlinespace[0.3em]
\multicolumn{2}{l}{\textbf{Ethnicity}}\\
\hspace{1em}Not Hispanic or Latina/o/x & 34 (82.9\%)\\
\hspace{1em}Hispanic or Latina/o/x & 3 (7.3\%)\\
\hspace{1em}I prefer not to answer & 3 \vphantom{1} (7.3\%)\\
\hspace{1em}Missing & 1 (2.4\%)\\
\addlinespace[0.3em]
\multicolumn{2}{l}{\textbf{Race}}\\
\hspace{1em}African American/Black & 3 (7.3\%)\\
\hspace{1em}Asian & 3 (7.3\%)\\
\hspace{1em}White & 27 (65.9\%)\\
\hspace{1em}More than one & 3 (7.3\%)\\
\hspace{1em}I prefer not to answer & 3 (7.3\%)\\
\hspace{1em}Missing & 2 (4.9\%)\\
\bottomrule
\end{tabular}
\caption{\label{demo-table}Participant demographics. All signers were exposed to ASL early in childhood. The dataset is not represented in racial, ethnic, and gender makeup.}
\label{table:demographics}
\end{table}

\subsection{Labelling}
We developed a novel method for labeling videos of signs which resolves some of the limitations of current methods using English glosses or phonological transcriptions as labels: we use videos of ASL signs as labels for ASL signs. The SignLab system presents the labeler with a video of a to-be-labeled sign and allows them to simultaneously search two lexical databases of ASL sign labels by typing in possible English translations (ASL-LEX and SignBank). The lexical databases were annotated to identify a variety of possible English translations for each sign, and all videos that had English translations that matched the typed input appeared in the search results. 
The labeler could visually scan the video thumbnails in the search results and play the videos by hovering their mouse over the thumbnail.
They could click to select an entry from the lexical databases that matched the production.
If both lexical databases contain the item, only the ASL-LEX label was presented to the labeler.
If the sign did not appear in either lexical database, the labeler could type in a free text description of the sign. 

With respect to lemmatizing, labelers were given the following instructions: \begin{itemize}
    \item If the sign and label mean the same thing, but look a little different (e.g., \href{https://asl-lex.org/visualization/?sign=duck}{DUCK} with two fingers versus four fingers): the sign and label match.
    \item If the sign and label mean the same thing, but look very different (e.g., \href{https://vimeo.com/378822204/9feba6bb0b}{CHILD} and \href{https://asl-lex.org/visualization/?sign=kid}{KID}): the sign and label do not match.
    \item Sign and labels that differ in more than one parameter (handshape, movement, or location) are probably not a match.
    \item If the sign and label mean something different, but look very similar (e.g., \href{https://asl-lex.org/visualization/?sign=peach}{PEACH} and \href{https://asl-lex.org/visualization/?sign=experience}{EXPERIENCE}): the sign and label do not match.
\end{itemize}

While labelers searched ASL-LEX by English translations, they were encouraged to ignore English when considering whether a sign was a match (e.g., “Do not worry if the English translation is not the one you would prefer to use. For example, if the ASL-LEX translation reads `father' and you prefer the English translation `dad,' just focus on whether the signs match). In some videos, participants mouthed English words while signing. Labelers could use English mouthing to the extent that it was helpful, and were free to match signs that differed in mouthing (e.g., a sign with the mouthing `dinner' could be a match to a reference video with the mouthing `supper'). If the labeler was unable to confidently label the sign, they marked it as uncertain, and these videos were excluded from the dataset (n = 2,288).

Before beginning to tag signs, labelers attended a training session with a member of the research team. They then independently tagged 100 training signs\footnote{These signs were randomly drawn from the dataset at the outset of labelling, and are not the same as the training fold of SemLex.} which were checked for inter-rater reliability with a set of correct answers developed by the research team. The team also examined responses for patterns of errors that reflected a misunderstanding of one or more of the training guidelines.  If the inter-rater reliability (Cohen's Kappa) was lower than .7, or if  systematic errors emerged when reviewing the training signs, we held another training meeting to review the responses and clarify the training guidelines before they proceeded. All labellers passed the .7 threshold after the second round of training signs. 

By labelling using lexical databases, the \name is cross-compatible with available linguistic resources for ASL, namely ASL-LEX \cite{sehyr2021asl, caselli2017asl}, ASL Citizen \cite{desai2023asl}, and the ASL SignBank \cite{hochgesang2019asl}. ASL-LEX contains detailed, manually annotated phonological descriptions of each of the 2,723 signs. 
These phonological transcriptions can be merged with the larger dataset as a ``broad transcription,'' making it possible to use phonological information in modeling without requiring manual annotation of the full dataset.
ASL SignBank has been used to label corpora of continuous signing \cite{slaaash}, which may also be leveraged in concert with the dataset we present here.

\section{Modeling Signs and Their Phonemes}
\label{method}
To provide empirical evidence that the \name data is both high-quality and practical, we conduct a suite of experiments related to sign and phoneme recognition.
The experiments are selected to answer a diverse array of research questions pertaining to sign and phoneme recognition:

\begin{itemize}
    \item[\ref{method:isr}] \textbf{Isolated sign recognition}: How accurate will a model be at recognizing isolated signs?
    
    \item[\ref{method:bpr}] \textbf{Phonological Feature Recognition}: How well will a model trained to recognize only the phonological features perform?
    
    \item[\ref{method:both}] \textbf{Phonological Feature+Isolated Sign Recognition}: How will a model benefit from learning signs in tandem with their phonological features?
    
    \item[\ref{method:test}] \textbf{Generalizability to Unseen \& Diverse Signers}: How sensitive is the model to spurious correlations among signers in the train set?
    
    \item[\ref{method:few-shot}] \textbf{Few-Shot Generalizability for ISR}: How well do models trained for Phonological Feature Recognition + ISR perform at recognizing signs with few training instances?
\end{itemize}

To answer these questions, we compare quantitative measures of performance (accuracy@k, mean reciprocal rank) across SL-GCN models (described below) learned on either WL-ASL or Sem-Lex training data for ISR and/or phonological feature recognition.

\subsection{The Sign Language Graph Convolution Network}
The SL-GCN model \cite{slgcn} is a specialized model for tasks involving sign language understanding.
It is an encoder-decoder model which takes a human pose estimation format of the input video and can be learned for one classification problem.
The SL-GCN encoder consists of ten repeated blocks, each of which contains (a) a decoupled GCN layer that encodes each keypoint in concert with its neighbors, (b) spatial and temporal attention over those keypoints, and (c) a temporal convolution layer.
The SL-GCN decoder consists of one fully-connected layer from the encoding to the desired output logits. 

We modify the decoder to allow for a variable number of classification heads by copying the encoding and providing it to multiple fully connected layers in parallel.
Structured this way, the SL-GCN model must encode all of the features that are pertinent to the classification tasks at hand in such a way that the decoder can easily separate the encoding into logits for each task.

This model architecture was selected for a variety of reasons. 
First, we use pose estimations over RGB video because it reduces not only the number of model parameters necessary to effectively process the input, but also the chance of biases due to spurious correlations between production and gender, race, or age.  
Second, the SL-GCN model contains separate attention mechanisms for space and time at each layer, improving the model's ability to recognize patterns over time (e.g. movement) or space (e.g. sign type).
And finally, there is empirical evidence that the SL-GCN model performs well on isolated sign recognition \cite{openhands}.

\subsection{Isolated Sign Recognition} \label{method:isr}
For the task of ISR, we use one classification head of size 2,731 (for the \name data) or 2,000 (for WLASL) coresponding to the number of target signs.
At the end of each forward pass, a cross-entropy loss is computed according to the one-hot encoding of the target label, and all model weights are trained while minimizing that loss.
We then compare the resulting accuracy (the correct answer is the top prediction), recall@$k$ (correct answer in the top-$k$ predictions), and mean reciprocal rank (1/rank of the correct answer) averaged across each item in the test set.

\subsection{Phonological Feature Recognition} \label{method:bpr}
For the task of phonological feature recognition, we train 16 classification heads ranging from size 2 to 58, one for each phonological feature type (see Table \ref{rw:tab:ptype} for the complete enumeration of types) that each take in the SL-GCN encoder representation of the sign video.
To compare with WLASL, we augment the dataset similarly to \citet{wlasl-lex} such that each video entry also contains estimations of its phonological features.
At the end of each forward pass, a \textit{summed} cross-entropy loss is computed according to the one-hot encoding of the target label within each type.
We then compare the resulting accuracy, recall-at-$k$, and mean reciprocal rank on the test set.

\subsection{Phonological Features + Sign Recognition} \label{method:both}
Following \citet{eacl}, we explore the possibility that ISR and phonological feature recognition are ``symbiotic'' tasks, meaning that a model which is trained to do both tasks simultaneously will be more accurate than one trained for either task alone.
We experiment with learning to recognize gloss alongside all 16 phonological feature types, as well as gloss alongside a small but informative subset of phonological feature types (handshape and minor location).
Otherwise, the model architecture is identical to the one described in Section \ref{method:bpr} only with an extra classification head for gloss.

\subsection{Generalizability to Unseen \& Diverse Signers} \label{method:test}
To explore the influence of spurious correlations between productions and the people who sign them (which is undesirable for most applications), we additionally compare the models trained for ISR and phonological feature recognition (separately) with regard to the validation set (seen and less diverse) vs. the test set (unseen and more diverse).
To the extent that the test set yields worse performance than the validation set, we may attribute some amount of the difference to the model relying on factors pertaining to race and/or gender.

\subsection{Few-Shot Generalizability for ISR} \label{method:few-shot}
To illustrate the practicality of learning phonology, we explore the average model performance with respect to the number of training instances per sign.
We compare the models described in Sections \ref{method:isr} and \ref{method:both} to provide empirical support that learning phonology enables a model to learn robust representations of signs more easily.
Among the itemized test results for each of these models, we first group signs by the number of instances found in training (in particular, those with 4--10 instances in the training set), and then compute the average performance within each group.

\section{Results}
\subsection{Isolated Sign Recognition}
When learned to recognize only gloss, the SL-GCN model has a top-1 accuracy of 67.7\%, a top-3 accuracy of 81.5\%, and a mean reciprocal rank (MRR) of 0.396 (see Table \ref{results:tab:wlasl_comparison}).
We juxtapose these results to WLASL, which has a smaller vocabulary of 2,000 signs, but the SL-GCN model performs worse, with a top-1 accuracy of 26.4\%, a top-3 accuracy of 45.7\%, and an MRR of 0.228.
This experiment shows that, relative to the WL-ASL benchmark, the \name data is well-labeled and therefore more tractible, but not trivial.

\begin{table}[t!]
    \centering
    
    \begin{tabular}{@{\extracolsep{8pt}}ccccccc@{}} 
        \toprule
        \multirow[c]{3}[3]{*}{\begin{tabular}{@{}c@{}} \textbf{Test Set} \end{tabular}} & \multicolumn{6}{c}{\textbf{Task}} \\
        \cmidrule{2-7}
        
         & \multicolumn{3}{c}{ISR} & \multicolumn{3}{c}{ISR+PFR} \\
        
        \cmidrule{2-4} \cmidrule{5-7}
        & $\textsc{acc}_1$ & $\textsc{acc}_3$ & \textsc{mrr} & $\textsc{acc}_1$ & $\textsc{acc}_3$ & \textsc{mrr} \\
        \midrule
        
        \texttt{WLASL-2000} & 26.4\% & 50.2\% & .43 & 38.1\% & 61.0\% & .52 \\

        \texttt{Sem-Lex} & 66.6\% & 81.5\% & .39 & 68.6\% & 82.0\% & .40 \\
        
        \bottomrule
    \end{tabular}
    \caption{Comparison of SL-GCN models trained with WLASL vs. Sem-Lex pose data ($\textsc{acc}_1 = \textit{top-1 accuracy}$, $\textsc{acc}_3 = \textit{top-3 accuracy}$, and $\textsc{mrr} = \textit{mean reciprocal rank}$). ISR models are trained to predict gloss only, ISR+PFR models predict both gloss and phonological features.}
    \label{results:tab:wlasl_comparison}
\end{table}

\begin{table}[t!]
    \centering
    \begin{tabular}{lcc} 
        \toprule
        \multicolumn{1}{c}{\multirow{2}{*}{\textbf{Phonological Feature Type}}} & \multicolumn{2}{c}{\textbf{Learning Method}} \\
        \multirow{2}{*}{} & Fine-Tune & Multitask \\
        \midrule
        Major Location                          & \textbf{0.877} & 0.875	\\
        \rowcolor{Gray} Minor Location          & \textbf{0.792} & 0.781	\\
        Second Minor Location                   & \textbf{0.787} & 0.772	\\
        \rowcolor{Gray} Contact                 & \textbf{0.893} & 0.886	\\
        Thumb Contact                           & \textbf{0.917} & 0.911	\\
        \rowcolor{Gray} Sign Type               & \textbf{0.889} & 0.879	\\
        Repeated Movement                       & \textbf{0.855} & 0.854	\\
        \rowcolor{Gray} Path Movement           & \textbf{0.756} & 0.754	\\
        Wrist Twist                             & 0.924 & \textbf{0.926}	\\
        \rowcolor{Gray} Selected Fingers        & \textbf{0.911} & 0.902 \\
        Thumb Position                          & \textbf{0.915} & \textbf{0.915}	\\
        \rowcolor{Gray} Flexion                 & \textbf{0.812} & 0.810	\\
        Spread                                  & \textbf{0.884} & 0.880	\\
        \rowcolor{Gray} Spread Change           & \textbf{0.903} & 0.895	\\
        Nondominant Handshape                   & \textbf{0.835} & 0.817 \\
        \rowcolor{Gray} Handshape               & \textbf{0.774} & 0.747 \\
        \midrule
        Average                                 & \textbf{0.858} & 0.850 \\
        \bottomrule
    \end{tabular}
    \caption{Phoneme feature recognition accuracy (top-1) between SL-GCN models fine-tuned to predict each type at a time or by learning them all at once, as evaluated on \sltest. All models are SL-GCNs pre-trained to predict gloss $y_g$ and then trained to predict phonological feature types $y_p$ ($p \in \mathcal{P}$) with the \sltrain dataset. Bold values indicate the highest per row.}
    \label{results:tab:phon}
\end{table}

\subsection{Phonological Feature Recognition}
Table \ref{results:tab:phon} shows the top-1 accuracies for phonological feature recognition (feature types described in Table \ref{rw:tab:ptype}). When learned to recognize the 16 phonological feature types presented in the \name, the SL-GCN is 85\% accurate on average regardless of how it learns them (individually by fine-tuning the entire model or by learning them all at once).
The most accurate phonological feature types were Wrist Twist (92.6\% accurate), Thumb Contact (91.7\% accurate), and Thumb Position (91.5\% accurate).
The least accurate types were Path Movement (75.6\% accurate), Handshape (77.4\% accurate), and Second Minor Location (78.7\% accurate).

\subsection{Phonological Features + Sign Recognition}
When learned to recognize both gloss and the 16 phonological feature types, the SL-GCN model is more accurate at ISR (71.3\%) than when trained to predict gloss alone (67.7\%).
This increase in performance is consistent with the results presented in \citet{eacl}, which shows that phonology is a useful auxiliary task to learning to recognize isolated signs.

\subsection{Few-Shot Generalizability}
Focusing on signs which are ``rare'' (i.e. had $4\leq n \leq 10$ examples during training), we observe a Pearson $r$ correlation of 0.73 between number of instances and average top-1 accuracy per sign class for \name.
This suggests a strong relationship between test accuracy and number of signs seen in training.
With only 4 signs in training, the SL-GCN model is able to recognize a sign with 62.2\% accuracy, and with 10 signs in training, that accuracy jumps to 72.3\%.
This is compared to WL-ASL, where the model recognizes 18.4\% and 31.3\%, respectively, for 4 and 10 training samples (see Table \ref{results:tab:gloss}).
Given the realistic, long-tailed distribution of signs in \name\ (specifically, 45\% signs have less than 10 instances), these findings indicate the SL-GCN model trained on \name\ is both effective at ISR, and in particular at recognizing signs with more consistent performance regardless of their frequency in the vocabulary.

Additionally, we report how learning gloss alongside phonological feature recognition influences few-shot generalizability.
The SL-GCN model, when learned to recognize both gloss and phonological features, is 68.2\% and 73.0\%, respectively, for 4 and 10 training samples.
In general, we observe that learning phonology as an auxiliary task not only improves overall gloss recognition accuracy, but also lessens the gap between less and more frequent signs.

\begin{table}[t!]
    \centering
    
    \begin{tabular}{@{\extracolsep{8pt}}cccccc@{}} 
        \toprule
        \multirow[c]{2}[2]{*}{\textbf{Dataset}} & \multirow[c]{2}[2]{*}{\textbf{Task}} & \multicolumn{4}{c}{\textbf{ Evaluation Set }} \\
        \cmidrule{3-6}

        & & $\texttt{val}_{all}$ & $\texttt{test}_{all}$ & $\texttt{test}_{n=10}$ & $\texttt{test}_{n=4}$ \\

        \midrule

        WLASL & ISR & --- & 26.4\% & 31.3\% & 18.4\% \\

        Sem-Lex & ISR & 68.2\% & 66.6\% & 72.3\% & 62.2\% \\

        Sem-Lex & ISR+PFR & \textbf{69.8\%} & \textbf{68.6\%} & \textbf{73.0\%} & \textbf{68.2\%} \\
        \bottomrule
    \end{tabular}
    \caption{Comparison* of ISR accuracy (top-1) for varying evaluation sets and learning targets. The validation set ($\texttt{val}_{all}$) and test set ($\texttt{test}_{all}$) intentionally differ with respect to signer race and gender, in addition to the latter set containing only unseen signers. $\texttt{test}_{n=k}$ is only the signs in the test set which have exactly $k$ corresponding instances in the training set. * Without zero-shot transfer from one test set to the other or human performance baselines, this comparison is limited in interpretability. }
    
    \label{results:tab:gloss}
\end{table}

\subsection{Seen vs. Unseen Signers}
In Table \ref{results:tab:gloss}, we additionally report the model's reliance on spurious correlations pertaining to individual signer differences by comparing performance on the validation set containing seen signers ($n=11,954$) and test set containing unseen signers representing more diverse demographics ($n=11,127$).
For seen signers, the SL-GCN trained to only predict gloss is 68.2\% accurate, while for unseen signers, the SL-GCN is 66.6\%.
These findings illustrate that there is a slight reliance on undesirable factors when learning to recognize signs.
Because we only use pose estimations of the videos, we believe the difference in performance is most likely attributable to differences in articulation, as opposed to visual differences among signers which are only observable with pixel-level information, such as skin color (which an RGB model might leverage to learn a spurious correlation with race or ethnicity).

\section{Discussion}
We present the \name\ for modeling ASL signs and their phonemes.
Our experiments show that Sem-Lex enables accurate models for recognizing signs and phonemes.
We additionally show that learning these tasks simultaneously improves accuracy across the board, including few-shot and unseen signers.
The success at few-shot generalization is especially true for the SL-GCN learned to predict both gloss and phonological features, demonstrating that learning phonology is an even more effective auxiliary task to learning ISR than previous work had shown.
However, there appears to be a slight reliance on spurious correlations, as demonstrated by the slightly lower performance on unseen and more diverse signers. A unique aspect of the \name is that the signs were spontaneously produced by deaf fluent signers using a widely-used experimental paradigm in psycholinguistic research. This approach ensures that signers were familiar with the signs they produced, and were not simply reproducing signs they may or may not know (e.g., \cite{desai2023asl}). 

\subsection{Limitations}
First, while there are more signs included in this benchmark than in other ASL datasets, it is still not representative of the full breadth of ASL. 
Our participants represent a small cross-section of all signers, who vary along many axes like experience and gender. The data is not representative of the larger population of ASL users in terms of race, ethnicity, and gender.
Additionally, fingerspelled words are underrepresented in the lexical databases we used for labelling, and so while participants may have contributed fingerspelled items, these are not among the labelled benchmark released here. 
Similarly, much of the morphology of ASL is not well represented in the labelled benchmark either (e.g., signs that are inflected for verb agreement, compound signs, etc.). 
\textit{Depicting signs} and \textit{classifier constructions}---semantically dense constructions which are unique to many signed languages---are also underrepresented in the \name.

Second, we note that models based on this benchmark alone (or any benchmark of isolated signs) may not generalize to continuous sign recognition (CSR).
By focusing on isolated signs, the benchmark is not representative of grammatical features (e.g. referential use of space, certain facial expressions) or coarticulation.
Researchers who intend to use these data or models for CSR or translation in any way should be aware of these discrepancies as they make and evaluate their models.

Finally, it should be noted that despite decades of sign linguistics research, many aspects of ASL phonology remain much less understood. 
The phonological descriptions of signs in ASL-LEX are incomplete, and so this paper represents an early step toward modeling sign phonology. 
While we did not conduct a direct validation of the models through research activities with the representative end users, this work is anchored in prior research involving the representative users and has been motivated by their priorities (see Section \ref{rw}).

\subsection{Accessing Data}
The goal of this paper is to share a benchmark which includes videos that were contributed with informed consent by deaf people who were compensated and recognized for their contributions (financially and/or via authorship). We hope that this benchmark is broadly useful, and spurs creativity and innovation. At the same time, ethical considerations for how sign language data are used are complex and sensitive \cite{bragg2021fate}. Prior to submitting this work, we convened a large group of deaf and signing scholars from a range of disciplines to consider how the community would like to share data. Following the recommendations of this group, we ask that users of these data:  
\begin{itemize}
    \item commit to ``do no harm,'' 
    \item work closely with deaf signing communities--the people who will be most impacted by sign language technology--to identify and mitigate possible harms, and maximize benefits to these communities
    \item recognize deaf contributors fairly (financially, through attribution, or other acknowledgement, as appropriate)
    \item work to mitigate possible power imbalances
    \item limit claims to those that are appropriate to the technology  (e.g., even high-performing ISR models do not obviate the need for human interpreters or teachers who are fluent in sign language)
\end{itemize}
We refer users who do not have connections to deaf communities to the \href{https://www.crest-network.com}{CREST} network at Gallaudet University, which aims to foster collaboration on sign-related technologies.

\subsection{Future Work}

The benchmark we present here was developed as part of a larger linguistic investigation of the semantic structure of the ASL lexicon. By identifying signs that people freely associate, we can learn how signs are related in meaning to one another. These associations can inform questions about how people learn and use signs. We are also eager to see this benchmark used for linguistic research (e.g., exploring variation in how different signers produce signs). 

Interdisciplinary work between linguists and technologists can be mutually beneficial. As we have laid out here, incorporating knowledge and resources from linguistics can aid in the development of sign language technology. Similarly, we believe modeling sign phonology will also benefit linguistics and psychology. Models of sign phonology can inform linguistic theories as to the phonological composition of signs. They can also be used to help build knowledge about relatively low-resource sign languages (e.g., those that do not have manually annotated databases), and can offer methods for cross-linguistic comparisons. This project paves the way for ethically sourced, efficient, and reproducible sign language research and more successful sign recognition technologies down the line.

\section{Conclusion}
The \name introduces new, high-quality data for modeling signs and their phonemes.
The 84,568 isolated sign productions were collected directly from Deaf participants with informed consent and financial compensation for their contributions.
Additionally, some 78\% are aligned with other datasets, allowing for phonological featurization for each video.
We show that modeling phonology is is worthwhile: when learned to classify phonological features in concert with gloss, a state-of-the-art model is able to recognize signs more accurately, and in particular signs that are rare.
With these data, we hope to inspire future work on studying signed languages in a more representative and ethical way, and with these insights, create more robust models for sign language understanding in direct collaboration with the Deaf community.

\bibliographystyle{ACM-Reference-Format}
\bibliography{sample-base}


\begin{thebibliography}{41}


\ifx \showCODEN    \undefined \def \showCODEN     #1{\unskip}     \fi
\ifx \showDOI      \undefined \def \showDOI       #1{#1}\fi
\ifx \showISBNx    \undefined \def \showISBNx     #1{\unskip}     \fi
\ifx \showISBNxiii \undefined \def \showISBNxiii  #1{\unskip}     \fi
\ifx \showISSN     \undefined \def \showISSN      #1{\unskip}     \fi
\ifx \showLCCN     \undefined \def \showLCCN      #1{\unskip}     \fi
\ifx \shownote     \undefined \def \shownote      #1{#1}          \fi
\ifx \showarticletitle \undefined \def \showarticletitle #1{#1}   \fi
\ifx \showURL      \undefined \def \showURL       {\relax}        \fi
\providecommand\bibfield[2]{#2}
\providecommand\bibinfo[2]{#2}
\providecommand\natexlab[1]{#1}
\providecommand\showeprint[2][]{arXiv:#2}

\bibitem[Athitsos et~al\mbox{.}(2008)]%
        {athitsos2008american}
\bibfield{author}{\bibinfo{person}{Vassilis Athitsos}, \bibinfo{person}{Carol
  Neidle}, \bibinfo{person}{Stan Sclaroff}, \bibinfo{person}{Joan Nash},
  \bibinfo{person}{Alexandra Stefan}, \bibinfo{person}{Quan Yuan}, {and}
  \bibinfo{person}{Ashwin Thangali}.} \bibinfo{year}{2008}\natexlab{}.
\newblock \showarticletitle{The american sign language lexicon video dataset}.
  In \bibinfo{booktitle}{\emph{2008 IEEE Computer Society Conference on
  Computer Vision and Pattern Recognition Workshops}}.
  \bibinfo{publisher}{IEEE}, \bibinfo{address}{Anchorage, AK, USA},
  \bibinfo{pages}{1--8}.
\newblock


\bibitem[Ball et~al\mbox{.}(2017)]%
        {ball2017history}
\bibfield{author}{\bibinfo{person}{Carolyn Ball} {et~al\mbox{.}}}
  \bibinfo{year}{2017}\natexlab{}.
\newblock \showarticletitle{The History of American Sign Language
  Interpreting}.
\newblock \bibinfo{journal}{\emph{Revue Internationale d'{\'E}tudes en Langues
  Modernes Appliqu{\'e}es}} \bibinfo{volume}{10}, \bibinfo{number}{Special}
  (\bibinfo{year}{2017}), \bibinfo{pages}{115--124}.
\newblock


\bibitem[Bragg et~al\mbox{.}(2021)]%
        {bragg2021fate}
\bibfield{author}{\bibinfo{person}{Danielle Bragg}, \bibinfo{person}{Naomi
  Caselli}, \bibinfo{person}{Julie~A Hochgesang}, \bibinfo{person}{Matt
  Huenerfauth}, \bibinfo{person}{Leah Katz-Hernandez}, \bibinfo{person}{Oscar
  Koller}, \bibinfo{person}{Raja Kushalnagar}, \bibinfo{person}{Christian
  Vogler}, {and} \bibinfo{person}{Richard~E Ladner}.}
  \bibinfo{year}{2021}\natexlab{}.
\newblock \showarticletitle{The fate landscape of sign language ai datasets: An
  interdisciplinary perspective}.
\newblock \bibinfo{journal}{\emph{ACM Transactions on Accessible Computing
  (TACCESS)}} \bibinfo{volume}{14}, \bibinfo{number}{2} (\bibinfo{year}{2021}),
  \bibinfo{pages}{1--45}.
\newblock


\bibitem[Bragg et~al\mbox{.}(2019)]%
        {bragg2019sign}
\bibfield{author}{\bibinfo{person}{Danielle Bragg}, \bibinfo{person}{Oscar
  Koller}, \bibinfo{person}{Mary Bellard}, \bibinfo{person}{Larwan Berke},
  \bibinfo{person}{Patrick Boudreault}, \bibinfo{person}{Annelies Braffort},
  \bibinfo{person}{Naomi Caselli}, \bibinfo{person}{Matt Huenerfauth},
  \bibinfo{person}{Hernisa Kacorri}, \bibinfo{person}{Tessa Verhoef},
  {et~al\mbox{.}}} \bibinfo{year}{2019}\natexlab{}.
\newblock \showarticletitle{Sign language recognition, generation, and
  translation: An interdisciplinary perspective}. In
  \bibinfo{booktitle}{\emph{Proceedings of the 21st International ACM SIGACCESS
  Conference on Computers and Accessibility}}. \bibinfo{pages}{16--31}.
\newblock


\bibitem[Brentari(1998)]%
        {brentari1998prosodic}
\bibfield{author}{\bibinfo{person}{Diane Brentari}.}
  \bibinfo{year}{1998}\natexlab{}.
\newblock \bibinfo{booktitle}{\emph{A prosodic model of sign language
  phonology}}.
\newblock \bibinfo{publisher}{MIT Press}, \bibinfo{address}{Cambridge, MA,
  USA}.
\newblock


\bibitem[Caselli et~al\mbox{.}(2017)]%
        {caselli2017asl}
\bibfield{author}{\bibinfo{person}{Naomi~K Caselli},
  \bibinfo{person}{Zed~Sevcikova Sehyr}, \bibinfo{person}{Ariel~M
  Cohen-Goldberg}, {and} \bibinfo{person}{Karen Emmorey}.}
  \bibinfo{year}{2017}\natexlab{}.
\newblock \showarticletitle{ASL-LEX: A lexical database of American Sign
  Language}.
\newblock \bibinfo{journal}{\emph{Behavior research methods}}
  \bibinfo{volume}{49} (\bibinfo{year}{2017}), \bibinfo{pages}{784--801}.
\newblock


\bibitem[Chen~Pichler and Hochgesang(nd)]%
        {slaaash}
\bibfield{author}{\bibinfo{person}{Deborah Chen~Pichler} {and}
  \bibinfo{person}{Julie Hochgesang}.} \bibinfo{year}{n.d.}\natexlab{}.
\newblock \bibinfo{booktitle}{\emph{Sign Language Acquisition, Annotation,
  Archiving and Sharing}}.
\newblock
\urldef\tempurl%
\url{https://slla.lab.uconn.edu/slaaash/}
\showURL{%
\tempurl}


\bibitem[De~Meulder et~al\mbox{.}(2019)]%
        {de2019legal}
\bibfield{author}{\bibinfo{person}{Maartje De~Meulder},
  \bibinfo{person}{Joseph~J Murray}, {and} \bibinfo{person}{Rachel~L McKee}.}
  \bibinfo{year}{2019}\natexlab{}.
\newblock \bibinfo{booktitle}{\emph{The legal recognition of sign languages:
  Advocacy and outcomes around the world}}.
\newblock \bibinfo{publisher}{Multilingual Matters}, \bibinfo{address}{Staple
  Hill, Bristol, UK}.
\newblock


\bibitem[Desai et~al\mbox{.}(2023)]%
        {desai2023asl}
\bibfield{author}{\bibinfo{person}{Aashaka Desai}, \bibinfo{person}{Lauren
  Berger}, \bibinfo{person}{Fyodor~O Minakov}, \bibinfo{person}{Vanessa Milan},
  \bibinfo{person}{Chinmay Singh}, \bibinfo{person}{Kriston Pumphrey},
  \bibinfo{person}{Richard~E Ladner}, \bibinfo{person}{Hal Daum{\'e}~III},
  \bibinfo{person}{Alex~X Lu}, \bibinfo{person}{Naomi Caselli},
  {et~al\mbox{.}}} \bibinfo{year}{2023}\natexlab{}.
\newblock \showarticletitle{ASL Citizen: A Community-Sourced Dataset for
  Advancing Isolated Sign Language Recognition}.
\newblock \bibinfo{journal}{\emph{arXiv preprint arXiv:2304.05934}}
  (\bibinfo{year}{2023}).
\newblock


\bibitem[Fenlon et~al\mbox{.}(2015)]%
        {fenlon2015building}
\bibfield{author}{\bibinfo{person}{Jordan Fenlon}, \bibinfo{person}{Kearsy
  Cormier}, {and} \bibinfo{person}{Adam Schembri}.}
  \bibinfo{year}{2015}\natexlab{}.
\newblock \showarticletitle{Building BSL SignBank: The lemma dilemma
  revisited}.
\newblock \bibinfo{journal}{\emph{International Journal of Lexicography}}
  \bibinfo{volume}{28}, \bibinfo{number}{2} (\bibinfo{year}{2015}),
  \bibinfo{pages}{169--206}.
\newblock


\bibitem[Hall et~al\mbox{.}(2019)]%
        {hall2019deaf}
\bibfield{author}{\bibinfo{person}{Matthew~L Hall}, \bibinfo{person}{Wyatte~C
  Hall}, {and} \bibinfo{person}{Naomi~K Caselli}.}
  \bibinfo{year}{2019}\natexlab{}.
\newblock \showarticletitle{Deaf children need language, not (just) speech}.
\newblock \bibinfo{journal}{\emph{First Language}} \bibinfo{volume}{39},
  \bibinfo{number}{4} (\bibinfo{year}{2019}), \bibinfo{pages}{367--395}.
\newblock


\bibitem[Hanke(2004)]%
        {hanke2004hamnosys}
\bibfield{author}{\bibinfo{person}{Thomas Hanke}.}
  \bibinfo{year}{2004}\natexlab{}.
\newblock \showarticletitle{HamNoSys-representing sign language data in
  language resources and language processing contexts}. In
  \bibinfo{booktitle}{\emph{LREC}}, Vol.~\bibinfo{volume}{4}.
  \bibinfo{pages}{1--6}.
\newblock


\bibitem[Hecht(2020)]%
        {hecht2020responsibility}
\bibfield{author}{\bibinfo{person}{Julia~L Hecht}.}
  \bibinfo{year}{2020}\natexlab{}.
\newblock \showarticletitle{Responsibility in the current epidemic of language
  deprivation (1990--present)}.
\newblock \bibinfo{journal}{\emph{Maternal and Child Health Journal}}
  \bibinfo{volume}{24}, \bibinfo{number}{11} (\bibinfo{year}{2020}),
  \bibinfo{pages}{1319--1322}.
\newblock


\bibitem[Hilger et~al\mbox{.}(2015)]%
        {hilger2015second}
\bibfield{author}{\bibinfo{person}{Allison~I Hilger},
  \bibinfo{person}{Torrey~MJ Loucks}, \bibinfo{person}{David Quinto-Pozos},
  {and} \bibinfo{person}{Matthew~WG Dye}.} \bibinfo{year}{2015}\natexlab{}.
\newblock \showarticletitle{Second language acquisition across modalities:
  Production variability in adult L2 learners of American Sign Language}.
\newblock \bibinfo{journal}{\emph{Second Language Research}}
  \bibinfo{volume}{31}, \bibinfo{number}{3} (\bibinfo{year}{2015}),
  \bibinfo{pages}{375--388}.
\newblock


\bibitem[Hill(2020)]%
        {hill2020deaf}
\bibfield{author}{\bibinfo{person}{Joseph Hill}.}
  \bibinfo{year}{2020}\natexlab{}.
\newblock \showarticletitle{Do deaf communities actually want sign language
  gloves?}
\newblock \bibinfo{journal}{\emph{Nature Electronics}} \bibinfo{volume}{3},
  \bibinfo{number}{9} (\bibinfo{year}{2020}), \bibinfo{pages}{512--513}.
\newblock


\bibitem[Hochgesang et~al\mbox{.}(2018)]%
        {hochgesang2018building}
\bibfield{author}{\bibinfo{person}{Julie Hochgesang}, \bibinfo{person}{OA
  Crasborn}, {and} \bibinfo{person}{Diane Lillo-Martin}.}
  \bibinfo{year}{2018}\natexlab{}.
\newblock \showarticletitle{Building the ASL Signbank. Lemmatization Principles
  for ASL}.
\newblock  (\bibinfo{year}{2018}).
\newblock


\bibitem[Hochgesang et~al\mbox{.}(2019)]%
        {hochgesang2019asl}
\bibfield{author}{\bibinfo{person}{Julie~A Hochgesang}, \bibinfo{person}{Onno
  Crasborn}, {and} \bibinfo{person}{Diane Lillo-Martin}.}
  \bibinfo{year}{2019}\natexlab{}.
\newblock \bibinfo{title}{ASL Signbank. New Haven, CT: Haskins Lab, Yale
  University}.
\newblock
\newblock


\bibitem[Jiang et~al\mbox{.}(2021)]%
        {slgcn}
\bibfield{author}{\bibinfo{person}{Songyao Jiang}, \bibinfo{person}{Bin Sun},
  \bibinfo{person}{Lichen Wang}, \bibinfo{person}{Yue Bai},
  \bibinfo{person}{Kunpeng Li}, {and} \bibinfo{person}{Yun~Raymond Fu}.}
  \bibinfo{year}{2021}\natexlab{}.
\newblock \showarticletitle{Skeleton Aware Multi-modal Sign Language
  Recognition}.
\newblock \bibinfo{journal}{\emph{2021 IEEE/CVF Conference on Computer Vision
  and Pattern Recognition Workshops (CVPRW)}} (\bibinfo{year}{2021}).
\newblock
\urldef\tempurl%
\url{https://openaccess.thecvf.com/content/CVPR2021W/ChaLearn/papers/Jiang_Skeleton_Aware_Multi-Modal_Sign_Language_Recognition_CVPRW_2021_paper.pdf}
\showURL{%
\tempurl}


\bibitem[Johnston and Schembri(1999)]%
        {johnston1999defining}
\bibfield{author}{\bibinfo{person}{Trevor Johnston} {and}
  \bibinfo{person}{Adam~C Schembri}.} \bibinfo{year}{1999}\natexlab{}.
\newblock \showarticletitle{On defining lexeme in a signed language}.
\newblock \bibinfo{journal}{\emph{Sign language \& linguistics}}
  \bibinfo{volume}{2}, \bibinfo{number}{2} (\bibinfo{year}{1999}),
  \bibinfo{pages}{115--185}.
\newblock


\bibitem[Joze and Koller(2018)]%
        {joze2018ms}
\bibfield{author}{\bibinfo{person}{Hamid Reza~Vaezi Joze} {and}
  \bibinfo{person}{Oscar Koller}.} \bibinfo{year}{2018}\natexlab{}.
\newblock \showarticletitle{MS-ASL: A large-scale data set and benchmark for
  understanding American Sign Language}.
\newblock \bibinfo{journal}{\emph{arXiv preprint arXiv:1812.01053}}
  (\bibinfo{year}{2018}).
\newblock


\bibitem[Kezar et~al\mbox{.}(2023)]%
        {eacl}
\bibfield{author}{\bibinfo{person}{Lee Kezar}, \bibinfo{person}{Jesse
  Thomason}, {and} \bibinfo{person}{Zed~Sevcikova Sehyr}.}
  \bibinfo{year}{2023}\natexlab{}.
\newblock \showarticletitle{Improving Sign Recognition with Phonology}. In
  \bibinfo{booktitle}{\emph{European Association for Computational
  Linguistics}}.
\newblock


\bibitem[Li et~al\mbox{.}(2020a)]%
        {li2020word}
\bibfield{author}{\bibinfo{person}{Dongxu Li}, \bibinfo{person}{Cristian
  Rodriguez}, \bibinfo{person}{Xin Yu}, {and} \bibinfo{person}{Hongdong Li}.}
  \bibinfo{year}{2020}\natexlab{a}.
\newblock \showarticletitle{Word-level deep sign language recognition from
  video: A new large-scale dataset and methods comparison}. In
  \bibinfo{booktitle}{\emph{Proceedings of the IEEE/CVF winter conference on
  applications of computer vision}}. \bibinfo{pages}{1459--1469}.
\newblock


\bibitem[Li et~al\mbox{.}(2020b)]%
        {wlasl}
\bibfield{author}{\bibinfo{person}{Dongxu Li}, \bibinfo{person}{Cristian
  Rodriguez}, \bibinfo{person}{Xin Yu}, {and} \bibinfo{person}{Hongdong Li}.}
  \bibinfo{year}{2020}\natexlab{b}.
\newblock \showarticletitle{Word-level Deep Sign Language Recognition from
  Video: A New Large-scale Dataset and Methods Comparison}. In
  \bibinfo{booktitle}{\emph{The IEEE Winter Conference on Applications of
  Computer Vision (WACV)}}.
\newblock


\bibitem[Marshall et~al\mbox{.}(2021)]%
        {marshall2021signed}
\bibfield{author}{\bibinfo{person}{Chloe Marshall}, \bibinfo{person}{Aurora
  Bel}, \bibinfo{person}{Sannah Gulamani}, {and} \bibinfo{person}{Gary
  Morgan}.} \bibinfo{year}{2021}\natexlab{}.
\newblock \showarticletitle{How are signed languages learned as second
  languages?}
\newblock \bibinfo{journal}{\emph{Language and Linguistics Compass}}
  \bibinfo{volume}{15}, \bibinfo{number}{1} (\bibinfo{year}{2021}),
  \bibinfo{pages}{e12403}.
\newblock


\bibitem[Mart{\'\i}nez et~al\mbox{.}(2002)]%
        {martinez2002purdue}
\bibfield{author}{\bibinfo{person}{Aleix~M Mart{\'\i}nez},
  \bibinfo{person}{Ronnie~B Wilbur}, \bibinfo{person}{Robin Shay}, {and}
  \bibinfo{person}{Avinash~C Kak}.} \bibinfo{year}{2002}\natexlab{}.
\newblock \showarticletitle{Purdue RVL-SLLL ASL database for automatic
  recognition of American Sign Language}. In
  \bibinfo{booktitle}{\emph{Proceedings. Fourth IEEE International Conference
  on Multimodal Interfaces}}. IEEE, \bibinfo{pages}{167--172}.
\newblock


\bibitem[Mesch and Wallin(2015)]%
        {mesch2015gloss}
\bibfield{author}{\bibinfo{person}{Johanna Mesch} {and} \bibinfo{person}{Lars
  Wallin}.} \bibinfo{year}{2015}\natexlab{}.
\newblock \showarticletitle{Gloss annotations in the Swedish Sign Language
  corpus}.
\newblock \bibinfo{journal}{\emph{International Journal of Corpus Linguistics}}
  \bibinfo{volume}{20}, \bibinfo{number}{1} (\bibinfo{year}{2015}),
  \bibinfo{pages}{102--120}.
\newblock


\bibitem[Neidle et~al\mbox{.}(2018)]%
        {Neidle2018NEWS}
\bibfield{author}{\bibinfo{person}{C. Neidle}, \bibinfo{person}{Augustine
  Opoku}, \bibinfo{person}{Greg Dimitriadis}, {and}
  \bibinfo{person}{Dimitris~N. Metaxas}.} \bibinfo{year}{2018}\natexlab{}.
\newblock \showarticletitle{NEW shared \& interconnected ASL resources:
  SignStream{\textregistered} 3 Software; DAI 2 for web access to
  linguistically annotated video corpora; and a sign bank}.
\newblock


\bibitem[of~America(nd)]%
        {lsa}
\bibfield{author}{\bibinfo{person}{Linguistic~Society of America}.}
  \bibinfo{year}{n.d.}\natexlab{}.
\newblock \bibinfo{booktitle}{\emph{Resolutions, Statements, Endorsements, and
  Related Actions}}.
\newblock
\urldef\tempurl%
\url{https://www.linguisticsociety.org/resource/resolutions-statements-and-guides}
\showURL{%
\tempurl}


\bibitem[of~the Deaf(nd)]%
        {wfd}
\bibfield{author}{\bibinfo{person}{World~Federation of~the Deaf}.}
  \bibinfo{year}{n.d}\natexlab{}.
\newblock \bibinfo{booktitle}{\emph{Human Rights of the Deaf}}.
\newblock
\urldef\tempurl%
\url{https://wfdeaf.org/our-work/human-rights-of-the-deaf/}
\showURL{%
\tempurl}


\bibitem[Pudans-Smith et~al\mbox{.}(2019)]%
        {pudans2019deaf}
\bibfield{author}{\bibinfo{person}{Kimberly~K Pudans-Smith},
  \bibinfo{person}{Katrina~R Cue}, \bibinfo{person}{Ju-Lee~A Wolsey}, {and}
  \bibinfo{person}{M~Diane Clark}.} \bibinfo{year}{2019}\natexlab{}.
\newblock \showarticletitle{To Deaf or not to deaf: That is the question}.
\newblock \bibinfo{journal}{\emph{Psychology}} \bibinfo{volume}{10},
  \bibinfo{number}{15} (\bibinfo{year}{2019}), \bibinfo{pages}{2091--2114}.
\newblock


\bibitem[Sandler(1987)]%
        {sandler1987sequentiality}
\bibfield{author}{\bibinfo{person}{Wendy Sandler}.}
  \bibinfo{year}{1987}\natexlab{}.
\newblock \bibinfo{booktitle}{\emph{Sequentiality and simultaneity in American
  Sign Language phonology}}.
\newblock \bibinfo{publisher}{The University of Texas at Austin}.
\newblock


\bibitem[Sch{\"u}ller et~al\mbox{.}(2021)]%
        {schuller2021lemma}
\bibfield{author}{\bibinfo{person}{Anique Sch{\"u}ller} {et~al\mbox{.}}}
  \bibinfo{year}{2021}\natexlab{}.
\newblock \showarticletitle{The Lemma Dilemma: Finding relevant lemmas to
  include in the Communicative Development Inventory for Sign Language of the
  Netherlands (NGT-CDI)}.
\newblock  (\bibinfo{year}{2021}).
\newblock


\bibitem[Sehyr et~al\mbox{.}(2022)]%
        {semassoc}
\bibfield{author}{\bibinfo{person}{S.~Z. Sehyr}, \bibinfo{person}{N. Caselli},
  \bibinfo{person}{A. Cohen-Goldberg}, {and} \bibinfo{person}{K. Emmorey}.}
  \bibinfo{year}{2022}\natexlab{}.
\newblock \showarticletitle{The semantic structure of American Sign Language:
  Evidence from free sign associations.}
\newblock \bibinfo{journal}{\emph{The 63rd Annual Meeting of The Psychonomic
  Society}} (\bibinfo{year}{2022}).
\newblock


\bibitem[Sehyr et~al\mbox{.}(2021)]%
        {sehyr2021asl}
\bibfield{author}{\bibinfo{person}{Zed~Sevcikova Sehyr}, \bibinfo{person}{Naomi
  Caselli}, \bibinfo{person}{Ariel~M Cohen-Goldberg}, {and}
  \bibinfo{person}{Karen Emmorey}.} \bibinfo{year}{2021}\natexlab{}.
\newblock \showarticletitle{The ASL-LEX 2.0 Project: A database of lexical and
  phonological properties for 2,723 signs in American Sign Language}.
\newblock \bibinfo{journal}{\emph{The Journal of Deaf Studies and Deaf
  Education}} \bibinfo{volume}{26}, \bibinfo{number}{2} (\bibinfo{year}{2021}),
  \bibinfo{pages}{263--277}.
\newblock


\bibitem[Selvaraj et~al\mbox{.}(2021)]%
        {openhands}
\bibfield{author}{\bibinfo{person}{Prem Selvaraj}, \bibinfo{person}{C.
  GokulN.}, \bibinfo{person}{Pratyush Kumar}, {and} \bibinfo{person}{Mitesh~M.
  Khapra}.} \bibinfo{year}{2021}\natexlab{}.
\newblock \showarticletitle{OpenHands: Making Sign Language Recognition
  Accessible with Pose-based Pretrained Models across Languages}. In
  \bibinfo{booktitle}{\emph{Annual Meeting of the Association for Computational
  Linguistics}}.
\newblock


\bibitem[Stokoe(1960)]%
        {StokoeWilliamC1960Slsa}
\bibfield{author}{\bibinfo{person}{William~C Stokoe}.}
  \bibinfo{year}{1960}\natexlab{}.
\newblock \bibinfo{booktitle}{\emph{Sign language structure: an outline of the
  visual communication systems of the American deaf.}}
\newblock \bibinfo{publisher}{Dept. of Anthropology and Linguistics, University
  of Buffalo}, \bibinfo{address}{Buffalo}.
\newblock
\showLCCN{61019055}


\bibitem[Tavella et~al\mbox{.}(2022)]%
        {wlasl-lex}
\bibfield{author}{\bibinfo{person}{Federico Tavella}, \bibinfo{person}{Viktor
  Schlegel}, \bibinfo{person}{Marta Romeo}, \bibinfo{person}{Aphrodite Galata},
  {and} \bibinfo{person}{Angelo Cangelosi}.} \bibinfo{year}{2022}\natexlab{}.
\newblock \showarticletitle{{WLASL}-{LEX}: a Dataset for Recognising
  Phonological Properties in {A}merican {S}ign {L}anguage}. In
  \bibinfo{booktitle}{\emph{Proceedings of the 60th Annual Meeting of the
  Association for Computational Linguistics (Volume 2: Short Papers)}}.
\newblock
\urldef\tempurl%
\url{https://aclanthology.org/2022.acl-short.49}
\showURL{%
\tempurl}


\bibitem[Van~der Kooij(2002)]%
        {van2002phonological}
\bibfield{author}{\bibinfo{person}{Els Van~der Kooij}.}
  \bibinfo{year}{2002}\natexlab{}.
\newblock \bibinfo{booktitle}{\emph{Phonological categories in Sign Language of
  the Netherlands: The role of phonetic implementation and iconicity}}.
\newblock \bibinfo{publisher}{Netherlands Graduate School of Linguistics}.
\newblock


\bibitem[Xiaoyan and Ruiling(2009)]%
        {xiaoyan2009survey}
\bibfield{author}{\bibinfo{person}{Xiao Xiaoyan} {and} \bibinfo{person}{Yu
  Ruiling}.} \bibinfo{year}{2009}\natexlab{}.
\newblock \showarticletitle{Survey on sign language interpreting in China}.
\newblock \bibinfo{journal}{\emph{Interpreting}} \bibinfo{volume}{11},
  \bibinfo{number}{2} (\bibinfo{year}{2009}), \bibinfo{pages}{137--163}.
\newblock


\bibitem[Yin et~al\mbox{.}(2021)]%
        {includingsl}
\bibfield{author}{\bibinfo{person}{Kayo Yin}, \bibinfo{person}{Amit Moryossef},
  \bibinfo{person}{Julie~A. Hochgesang}, \bibinfo{person}{Yoav Goldberg}, {and}
  \bibinfo{person}{Malihe Alikhani}.} \bibinfo{year}{2021}\natexlab{}.
\newblock \showarticletitle{Including Signed Languages in Natural Language
  Processing}. In \bibinfo{booktitle}{\emph{Association for Computational
  Linguistics (ACL)}}.
\newblock


\bibitem[Zahedi et~al\mbox{.}(2005)]%
        {zahedi2005combination}
\bibfield{author}{\bibinfo{person}{Morteza Zahedi}, \bibinfo{person}{Daniel
  Keysers}, \bibinfo{person}{Thomas Deselaers}, {and} \bibinfo{person}{Hermann
  Ney}.} \bibinfo{year}{2005}\natexlab{}.
\newblock \showarticletitle{Combination of tangent distance and an image
  distortion model for appearance-based sign language recognition}. In
  \bibinfo{booktitle}{\emph{Pattern Recognition: 27th DAGM Symposium, Vienna,
  Austria, August 31-September 2, 2005. Proceedings 27}}. Springer,
  \bibinfo{pages}{401--408}.
\newblock


\end{thebibliography}

\appendix

\end{document}